\documentclass{article}
 \pdfoutput=1

     \usepackage[final, nonatbib]{neurips_2019}

\usepackage[utf8]{inputenc} 
\usepackage[T1]{fontenc}    
\usepackage{hyperref}       
\usepackage{url}            
\usepackage{booktabs}       
\usepackage{amsfonts}       
\usepackage{nicefrac}       
\usepackage{microtype}      

\usepackage{amsmath}
\usepackage{amssymb}
\newcommand{\N}{\mathbb{N}}
\newcommand{\A}{\mathbb{A}}

\usepackage{graphicx}
\usepackage{caption}

\usepackage{makecell}

\title{Exploring Graph Neural Networks for Stock Market Predictions with Rolling Window Analysis}

\author{%
  Daiki Matsunaga \thanks{Equal contribution.}\\
  IBM Japan\\
  19-21 Nihonbashi, Hakozaki-cho, Chuo-ku\\
  Tokyo, Japan \\
  \texttt{e36957@jp.ibm.com} \\
   \And
   Toyotaro Suzumura \textsuperscript{*}\\
   IBM T.J. Watson Research Center \\
   1101 Kitchawan Rd, Yorktown Heights\\
   New York, USA \\
   \texttt{tsuzumura@us.ibm.com} \\
   \AND
   Toshihiro Takahashi \\
   IBM Research - Tokyo \\
   19-21 Nihonbashi, Hakozaki-cho, Chuo-ku\\
   Tokyo, Japan \\
   \texttt{e30137@jp.ibm.com} \\
}

\begin{document}

\maketitle

\begin{abstract} 
Recently, there has been a surge of interest in the use of machine learning to help aid in the accurate predictions of financial markets. Despite the exciting advances in this cross-section of finance and AI, many of the current approaches are limited to using technical analysis to capture historical trends of each stock price and thus limited to certain experimental setups to obtain good prediction results. On the other hand, professional investors additionally use their rich knowledge of inter-market and inter-company relations to map the connectivity of companies and events, and use this map to make better market predictions. For instance, they would predict the movement of a certain company's stock price based not only on its former stock price trends but also on the performance of its suppliers or customers, the overall industry, macroeconomic factors and trade policies. This paper investigates the effectiveness of work at the intersection of market predictions and graph neural networks, which hold the potential to mimic the ways in which investors make decisions by incorporating company knowledge graphs directly into the predictive model. The main goal of this work is to test the validity of this approach across different markets and longer time horizons for backtesting using rolling window analysis. In this work, we concentrate on the prediction of individual stock prices in the Japanese Nikkei 225 market over a period of roughly 20 years. For the knowledge graph, we use the Nikkei Value Search data, which is a rich dataset showing mainly supplier relations among Japanese and foreign companies. Our preliminary results show a 29.5\% increase and a 2.2-fold increase in the return ratio and Sharpe ratio, respectively, when compared to the market benchmark, as well as a 6.32\% increase and 1.3-fold increase, respectively, compared to the baseline LSTM model.
\end{abstract}

\section{Introduction}\label{intro}

\subsection{Traditional investment strategies}
Traditionally, investors and analysts in the finance industry have focused on two major approaches in predicting the movement of stock prices: fundamental analysis and technical analysis \cite{beyaz}. The former focuses on indicators which attempt to quantify the intrinsic value of a company, such as macroeconomic analysis (e.g. Gross Domestic Product, Consumer Price Index), industry analysis (i.e. industry-wide trends) and company analysis (e.g. Price-to-Earnings ratio) in order to predict the growth potential of the company in question. On the other hand, technical analysis focuses on various price movement indicators such as momentum, volatility and moving averages, all of which rely upon past trends as future indicators for each respective company \cite{shah}. This latter approach presupposes that all of the necessary information regarding each stock in question are included in its stock price. The difference in these two approaches stem from the longstanding debate about whether or not stock markets are inherently predictable and thus profitable, either in the short-term or in the long-term. However, there is a commonality in both approaches in that they use historical indicators to predict each individual stock in question, which is especially the case for most machine learning approaches. Indeed, much of the work in machine learning fits into the category of technical analysis as it is a natural way to formulate stock price prediction as a sequential modeling task \cite{shah}. 

\subsection{An increasingly interconnected financial world}
In reality, however, the world is increasingly more interconnected through the proliferation of the Internet, the rise of Fintech and global banking conglomerates operating across borders. In this complex financial world, stock prices, market indices, and other crucial indicators work and move in tandem, whereby a multitude of political and economic factors can affect each price. We need to look no further than the 2008 financial crisis to understand this interconnectivity intuitively and understand that individual prices don’t move solely from past trends in isolation \cite{oecd}. As a sidenote, this phenomena makes it increasingly more difficult to make sound investment decisions based solely on past trends from a few predictors, and thus more susceptible to risky investments or even fraud on a massive scale. For instance, the case of Valeant Pharmaceuticals 
is a good example of a massive fraud case which appeared to have been a sound investment choice if focusing on its linearly soaring stock price trend without understanding the full context of its suspicious business model of acquiring many drug companies and racking up their drug prices for profit. It was a very popular and attractive choice among investors, before it lost north of 60\% of its original stock value in 2015, leading to huge losses for both individual and institutional investors. Although this case mostly highlights the malintent of some investors and executives, it also shows that the financial world is quickly becoming too complex for any one person to have a holistic view of the market or dive deeper into the multitude of factors that affect a particular index or stock price \cite{grove2017forensic}.

\subsection{Graph-based machine learning in the new era of finance}
From a machine learning perspective, network structures such as company knowledge graphs were difficult to incorporate due to its non-Euclidean nature. Thus, most statistical approaches, even with recent deep learning methods, did not incorporate interconnectivity into the predictive model.  Furthermore, graph analytics traditionally involved hand-engineering features and summary graph statistics using graph algorithms such as PageRank, egonet, degree distribution, among others \cite{aml}\cite{replearning}. In stock market prediction, technical analysis falls under this umbrella whereby some of the aforementioned indicators such as moving averages are hand-engineered from raw price data and fed into a classifier as features. This approach, however, is both time-consuming and inflexible, especially in representing the dynamic nature of global financial markets.
One possible remedy to these issues is the recent innovation in graph neural network algorithms.  
These algorithms present an opportunity to make predictive models which better represent the increasingly fast-paced and interconnected nature of the global financial system by incorporating  knowledge graphs. This type of research frames the feature extraction process as a learning task rather than a preprocessing step, as is the case with graph summary statistics approaches. We posit that the financial system can best be represented in this network structure where companies are connected through various relation types such as supplier/customer, shareholder and industries. Similar to how a professional investor makes decisions, graph neural networks can utilize the network structure to incorporate the interconnectivity of the market and make better stock price predictions, rather than relying solely on the historical stock prices of each individual company or on hand-crafted features. Despite its promise, however, the use of these algorithms to the stock market prediction problem has only recently started to be explored. 

The key contribution of this paper is as follows.
\begin{itemize}

\item We empirically evaluate the effectiveness of graph neural networks on stocks listed in the Nikkei 225 market.
\item We conduct backtesting over a period of roughly 20 years (4,632 timesteps) to test the generalizability of graph neural networks as a method for predicting stock prices
\item We conduct backtesting using rolling window analysis, which is an effective method for evaluating time series data, especially for historical stock price data.
\end{itemize}

The remainder of the paper is organized as follows. In Section \ref{survey}, we review related work in the use of machine learning approaches to making stock market predictions. In section \ref{proposed}, we detail the algorithm and methodology used in our work. In Section \ref{evaluation} we outline the datasets used, evaluation methods and the analysis of preliminary results. We finally discuss some future extensions of this work in Section \ref{discussion} and conclude in Section \ref{conclusion}.

\section{Related work} \label{survey}
\subsection{Deep learning for technical analysis}
There have been a number of works in the use of statistical and machine learning methods for analyzing past stock trends. This problem can be framed as either a classification task e.g. UP, DOWN, STAY, or a regression task e.g. prediction of next day's closing price \cite{beyaz}. A number of approaches \cite{hu2018listening} \cite{dlframework2017bao} have shown that deep learning methods outperform traditional machine learning approaches which use hand-crafted features. \cite{dlframework2017bao} combines autoencoders and Long-Short Term Memory (LSTM) networks in order to predict the indices of six different markets across the world and has shown its predictive capabilities across various developing and developed markets. The State Frequency Memory model in \cite{hu2018listening} extends the LSTM model with inspiration from the signal processing community whereby the model attempts to decompose stock price trends into multiple latent frequency components. This approach outperforms linear Autoregressive (AR) models in the prediction of 50 stocks over a period of 9 years. Although the architectural details of the models vary, they show that deep learning approaches are effective in finding non-linear patterns in the stock market.

\subsection{Natural language processing approaches}
The use of natural language processing (NLP) approaches have been investigated as another interesting way to incorporate natural language text such as SEC filings \cite{lee-etal-2014-importance}, patents \cite{patent} or financial news \cite{ATKINS2018120}. Features such as n-grams can be extracted \cite{lee-etal-2014-importance}, or word embedding approaches such as Word2vec  can be used as an unsupervised feature extractor \cite{hu2018listening}. The Hybrid Attention Network model in \cite{hu2018listening} uses Word2vec to extract embeddings for each word in news articles, and aggregates the embeddings to output a fixed size vector that represents the features for each news article. These features can then be fed into a supervised classifier similar to how numerical data would be used as inputs.

\subsection{Complex networks and network science}
A related area of research in stock prediction using graphs comes from the network science community, especially with the study of complex networks. Complex networks study a specific type of random graphs which can broadly be categorized as small-world or scale-free. Small-world networks randomly connect two distant nodes, shortening the diameter of a graph. Scale-free networks follow a power-law connectivity distribution, where a small number of hub nodes have a large number of connections \cite{supplychain2013}. In the stock prediction setting, complex networks are used to study the major influencer companies (network centrality) or clustered communities (modularity), without specifying any explicit edge relationships. For example, \cite{DBLP:journals/corr/KimS17} makes random graphs at each time interval by connecting companies through the mutual information of price movements, in order to predict the S\&P 500 index. For each random graph, they extract graph information such as the node strength distributions, network centrality and modularity. They show that incorporating these graph characteristics can improve upon ARIMA models, in predicting the one-minute interval fluctuations of the S\&P 500 market index. 

\subsection{Company knowledge graphs for market prediction}
Recently, there have been interesting works \cite{feng2019temporal}\cite{kim2019hats} which incorporate open source knowledge graphs (e.g.Wikidata) into their stock prediction models for individual companies by combining graph neural networks and sequential models such as Recurrent Neural Networks (RNN). Both works follow four major steps with some differences in the algorithmic details.
First, historical stock price data are fed into an RNN layer as features to output node embeddings for each stock in question. The features can be a combination of the previous day's closing price and moving averages \cite{feng2019temporal} or price change rates \cite{kim2019hats}. These features are used to capture the short term and/or mid term trend of each stock price. Second, an adjacency matrix is created through open knowledge graph sources such as Wikidata, to connect companies through industry, subsidiary, people and product relations. Third, the node (company) embeddings created by the RNN layer and the adjacency matrix are combined and fed into a graph neural network layer to update the node embeddings. Finally, the updated embeddings are fed into a fully connected layer to make price predictions. Furthermore, \cite{kim2019hats} feeds the updated node embeddings into a separate graph pooling layer to predict the movement of market indices e.g. S\&P 500 Index, making the overall architecture a multi-task one.
We revisit the algorithm in \cite{feng2019temporal} in more detail in Section \ref{proposed}.

\section{Generalizability of GNNs for stock prediction} \label{proposed}
In this paper we examine the following research question: \\
\textit{Are knowledge graph datasets e.g. company relations data, useful in making stock predictions across different markets and longer time horizons? } \\
To answer this question, we have adopted and extended the work in \cite{feng2019temporal} and the corresponding open source codebase. 

\subsection{Graph convolutional network}
The basis for the work in \cite{feng2019temporal} is the Graph Convolutional Network (GCN) algorithm proposed in \cite{kipf}. The algorithm requires as input (a) features for each node and (b) an adjacency matrix showing relations among nodes. The features for each node \begin{math}i\end{math} are updated by using the given adjacency matrix to aggregate feature representations of its neighbors: \\

\begin{equation}
    \label{equation:gcn_}
    \displaystyle \overline{e_i} = \sum_{j}^{N} (\frac{\phi\ (W^T\A_{ij} + b)}{d_j} \times e_j)
\end{equation}

where for \begin{math}N \end{math} companies and \begin{math}K \end{math} relation types,
\begin{math} \A \in \mathbb{R}^{\N \times \N \times K}\end{math} is the adjacecy matrix,
\begin{math}d_j \end{math} is the degree of node \begin{math}j \end{math} acting as a normalization factor, 
\begin{math} W \in \mathbb{R}^{K}\end{math} and  \begin{math}b \end{math} are learnable parameters 
and \begin{math}\phi \end{math} is a non-linear activation. 
 Along the diagonal of the adjacency matrix, \begin{math}\A_{ii} \end{math} always equals 1 to ensure that a node's own embeddings are being aggregated as well. The updated embeddings can be fed into either additional GCN layers for further neighborhood sampling or into a fully connected layer for the specific learning task such as classification or regression. For the stock market prediction setting, the features for node \begin{math}e_i \end{math} can be the output of an RNN layer, where the sequential data is mapped into a vector of fixed length i.e.
\begin{equation}
    \label{equation:features}
    \displaystyle e_i = RNN(\chi_i)
\end{equation}
where \begin{math} \chi_i \end{math} is the features calculated from historical stock price data such as closing price or moving averages. Equation \ref{equation:features} is calculated for each relevant stock at each timestep.
In this work, we follow the algorithm in \cite{feng2019temporal} and use an LSTM layer rather than the vanilla RNN layer. As for hyperparameters, we followed the hyperparameters used in \cite{feng2019temporal} to train on NYSE stock prices, and set the number of LSTM units at 32 and the sequence length at 8 timesteps.
\subsection{Temporal graph convolution}
The implementation of GCNs, while able to incorporate graph data into the predictive model, only uses a static version of the adjacency matrix. In other words, the weighing factor which updates \begin{math}e_{i} \end{math} to \begin{math}\overline{e_i} \end{math} is static across time. In reality, however, the propagation of stock trends among connected companies evolves dynamically. For instance, the stock price between car manufacturer Company X and car parts supplier Company Y might be highly correlated as a default, but there could be instances where, for example, a new car model release could strengthen this relationship to an even greater degree.
To incorporate this time-sensitivity into the relations among nodes (companies), the dot product of the embeddings for nodes \begin{math}i \end{math} and \begin{math}j \end{math} are multiplied as a weighing factor for the relationship strength.

\begin{equation}
    \label{equation:gcn}
    \displaystyle \overline{e_i} = \sum_{j}^{N}( e_i^T\ e_j\ \times \frac{\phi\ (W^T\A_{ij} + b)}{d_j}\ \times e_j)
\end{equation}

Intuitively, the dot product represents the similarity between the embbeddings of nodes \begin{math}i \end{math} and \begin{math}j \end{math}; the more similar the recent trends in stock prices, the stronger the relation strength will be. Again, equation \ref{equation:gcn} is calculated for each relevant company at each timestep. These embeddings are then fed into a fully-connected layer for regression.

As one minor adjustment, as opposed to the implementation for the paper optimizing the RNN layers and TGC layers separately, we've instead optimized the whole network in an end-to-end manner. This is to avoid unnecessary bugs and speed up training \cite{glasmachers2017limits}, especially in a relatively simple network with one RNN layer, one TGC layer and one fully connected layer.

\subsection{Rolling window analysis}
In the financial sector, professionals are often concerned with models which reliably can predict the market over long time horizons. Therefore, we have implemented the rolling window analysis method \cite{zivot2006rolling} for splitting the full dataset into the training and test set. We use 2000 timesteps for training and 200 for testing as a fixed window, and slide that window over the full number of timesteps. For example, index 0 to 2000 and 2000 to 2200 will be used in the first iteration for training and testing, respectively. The second iteration will have indices 200 to 2200 and 2200 to 2400, and so on. This method not only ensures that the model is generalizeable across long time horizons but also represents the idea that recent stock prices have better predictive power than older ones. For instance, in order to make future predictions for a particular stock, it makes more intuitive sense to use the historical prices from the last five years rather than from the 1980s. As a further note, it is also possible to implement the growing window method, whereby the window size for the training set continues to increase while the test set window continues to slide. For instance, the first iteration contains indices 0 to 2000 as the training set, the second iteration contains 0 to 2200, and so on. This method can be useful especially in cases where the data is sparse with wider time intervals or if there are difficulties in obtaining a comprehensive number of timesteps for backtesting.
\section{Evaluation} \label{evaluation}

\subsection{Historical stock price data}
We test our model on Japanese companies listed in the Nikkei 225 market. The number of companies, however, is filtered down to 176 from 225 listed companies. This is to ensure that there are enough timesteps to conduct long-term backtesting, since some companies were either founded or listed only recently and thus contains fewer timesteps. For features to be fed into the LSTM layer, we follow \cite{feng2019temporal} and calculate 5, 10, 20, 30-day moving averages at each timestep in addition to the adjusted closing prices. We also calculate the 1-day return ratio at timestep \begin{math} t, R_t = \frac{p_t\ -\ p_{t-1}}{p_{t-1}} \end{math} as the ground truth, where \begin{math}p_t \end{math} represents the price at timestep \begin{math}t \end{math}. 

\begin{table}
  \caption{An example company relation data for a particular Company X}
  \label{company-x}
  \centering
  \begin{tabular}{ccccc}
    \toprule
    \cmidrule(r){1-2}
    Relation Type     & Company Name     & Sector  & Industry & Country\\
    \midrule
    Supplier & Company A  & \makecell{Producer  Manufacturing} & Auto Parts: OEM & JPN     \\
    Customer     & Company B & Finance & Finance/Retail/Leasing & USA     \\
    Partner     & Company C       & \makecell{Consumer Durables}& Motor Vehicles & USA  \\
    \bottomrule
  \end{tabular}
\end{table}

\begin{table}
  \caption{Knowledge graph summary}
  \label{graph-info}
  \centering
  \begin{tabular}{ll}
    \toprule
    \# of Nodes & 12, 473     \\
    \# of Edges & 38, 252     \\
    Average Node Degree & 6.13    \\
    \bottomrule
  \end{tabular}
\end{table}

\begin{table}
  \caption{First- and second-order relations extracted from the created company knowledge graph}
  \label{relation-types}
  \centering
  \begin{tabular}{lll}
    \toprule
    \cmidrule(r){1-2}
    Relation Type     & \# of edges     & Description  \\
    \midrule
    supplies-from & 145 & First-order relation between companies connected by Supplier relation\\
    customer-of & 145 & First-order relation between companies connected by Customer relation\\
    partner-with & 555 & First-order relation between companies connected by Partner relation\\
    shares-owned-by & 220 & First-order relation between companies connected by Shareholder relation\\
    common-industry & 438 & Second-order relation between companies with common industry \\
    common-customer & 3634 & Second-order relation between companies with common customers \\
    common-supplier & 8262 & Second-order relation between companies with common supplier \\
    \bottomrule
  \end{tabular}
\end{table}

\subsection{Knowledge graph dataset}
 We used the Nikkei Value Search dataset 
, which consists of supplier, customer, partner and shareholder relations for each of the 176 relevant companies, as shown in Table \ref{company-x}. This data is then used to create a comprehensive knowledge graph which includes companies outside of Nikkei 225 index as well as Japan. The summary statistics for this comprehensive knowledge graph is shown in Table \ref{graph-info}.  
The nodes have a "node\_type" attribute with values "company", "sector", "industry" or "country", while edges have an "edge\_type" attribute with values "in-country",
"in-industry",
"in-sector",
"parent-company-of",
"partner-with",
"related-to",
"same-company", or
"shareholder". 
The nodes also have a boolean attribute called "in\_nikkei" to distinguish the nodes which are listed in the Nikkei 225 market. Also, "same-company" relations were used for entity resolution purposes, where English and Japanese names for the same company were concatenated as identical nodes.
These preprocessing steps allow for the extraction of first-order and second-order relations, listed in Table \ref{relation-types}. For example, if two nodes listed in the Nikkei 225 market both had an"in-industry" edge connected to a common "industry" node, the two nodes can be connected via second-order connections. 

First-order relations are direct relations among the 176 companies. This shows an apparent relationship where connected stocks could be related. However, these relationships are quite sparse and also an area where investors are already actively using for analyzing individual stocks. To alleviate the sparsity problem and make better use of the capabilities of graph neural networks, we've extracted second-order relations, which are indirect connections between any two of the 176 companies in question. The result of the preprocessing step are seven different types of adjacency matrices, \begin{math} \A \in \mathbb{R}^{N \times N } \end{math} where \begin{math}N\end{math} is the number of companies. We've also tested a relation type called "all" where we stack the seven adjacency matrices into a three dimensional matrix, \begin{math} \A \in \mathbb{R}^{\N \times \N \times K} \end{math} where \begin{math}K\end{math} is the number of relation types. 

\subsection{Evaluation Metrics} 
As an evaluation metric, we use the return ratio and the Sharpe ratio, as they are both used frequently in the finance industry. The return ratio \begin{math}R_t\end{math} is calculated as \begin{math}  \frac{p_t - p_{t-1}}{p_{t-1}} \end{math}, which is simply the percentage increase in the stock price or the return when we buy a stock at timestep \begin{math}  t-1 \end{math} and sell at timestep \begin{math}  t \end{math}. The Sharpe ratio is calculated as
\begin{math}  \frac{R_t - R_f}{\sigma} \end{math}
where the return \begin{math}R_t\end{math} is subtracted by the risk-free rate \begin{math}R_f\end{math} which is then divided by the standard deviation of the returns \begin{math}\sigma\end{math}. The standard deviation \begin{math}\sigma\end{math} represents the risk associated with a stock, which is important in avoiding the evaluation of models based purely on returns. It is important to note that the Sharpe Ratio should not be evaluated as an absolute value, but rather viewed as a performance indicator in comparison with the benchmark.   

\begin{table}
  \caption{Performance comparison among various graph types against LSTM and Buy-Hold benchmarks}
  \label{final-results}
  \centering
  \begin{tabular}{lll}
    \toprule
    \cmidrule(r){1-2}
    Edge Type & Return Ratio (Yearly Rates) & Sharpe Ratio\\
    \midrule
    Benchmark (Buy-Hold) & 4.30\% & 0.16 \\
    LSTM & 23.63\% & 0.29 \\
    supplier-of & 21.10\% & 0.30 \\
    customer-of & 27.52\% & 0.35 \\
    partner-with & 15.48\% & 0.12 \\
    shares-owned-by & 23.54\% & 0.22 \\
    common-industry & 18.09\% & 0.24 \\
    common-customer & 18.61\% & 0.17 \\
    common-supplier & 16.27\% & 0.14 \\
    \textbf{all} & \textbf{29.95\%} & \textbf{0.37} \\
    
    \bottomrule
  \end{tabular}
\end{table}

\begin{table}
  \caption{Return ratio for the prediction of various timesteps (customer-of)}
  \label{results-timesteps}
  \centering
  \begin{tabular}{ll}
    \toprule
    \cmidrule(r){1-2}
    \# of time steps & Return Ratio (Yearly Rates)\\
    \midrule
    \textbf{1-day} & \textbf{27.52\%}  \\
    5-day & 21.93\%  \\
    10-day & 23.327\%  \\
    20-day & 17.36\%  \\
    \bottomrule
  \end{tabular}
\end{table}

\subsection{Evaluation Results}
Our preliminary results in Table \ref{final-results} shows the following results.

(a) Both the LSTM model and the various graph neural network models outperform the market benchmark.

(b) The LSTM model is outperformed by graph neural networks for "customer-of" and "all" in terms of both the Sharpe and return ratio, and is outperformed by "supplier-of" in terms of the Sharpe ratio.

(c) "all" relations have the best performance based on both the return ratio and Sharpe ratio.

(d) "customer-of" relations had a relatively high return and Sharpe ratio despite its sparsity, which suggests that customer relations are highly effective in improving predictive performance.

The results for (d) makes intuitive sense since supply chain analysis is an important industry practice among investors. However, it is important to note that while customer relations seemed to be effective in the aggregate, its effectiveness varies by time periods. For example, \cite{feng2019temporal} concluded that partner relationships e.g. product collaborations were the most effective relation type for the NASDAQ market. As the authors mention, it does make intuitive sense for companies who collaborated on a common product to have connected revenue. However, the low performance of the partner relations in this study suggest three possible theories. First, since the testing period for the paper in \cite{feng2019temporal} was limited to 200 timesteps between 2017 and 2018, the results were inconclusive. Second,  partner relations could indeed have high predictive power in the NASDAQ market even with extended backtesting, but not generalizable to the Nikkei 225 market. Finally, it could also be the case that since work in \cite{feng2019temporal} used a Wikipedia-based knowledge graph, the obtained partner relationships were filtered to very important or well-known ones, while the Nikkei Value Search dataset was more comprehensive.

\subsection{Qualitative Analysis}
\begin{figure}
    \centering
    \caption{ Comparison of stock price movements between Company A and its customer: Company B. }
    \label{fig:customer}
   
    \begin{minipage}{0.45\textwidth}
        \centering
        \includegraphics[width=0.9\textwidth]{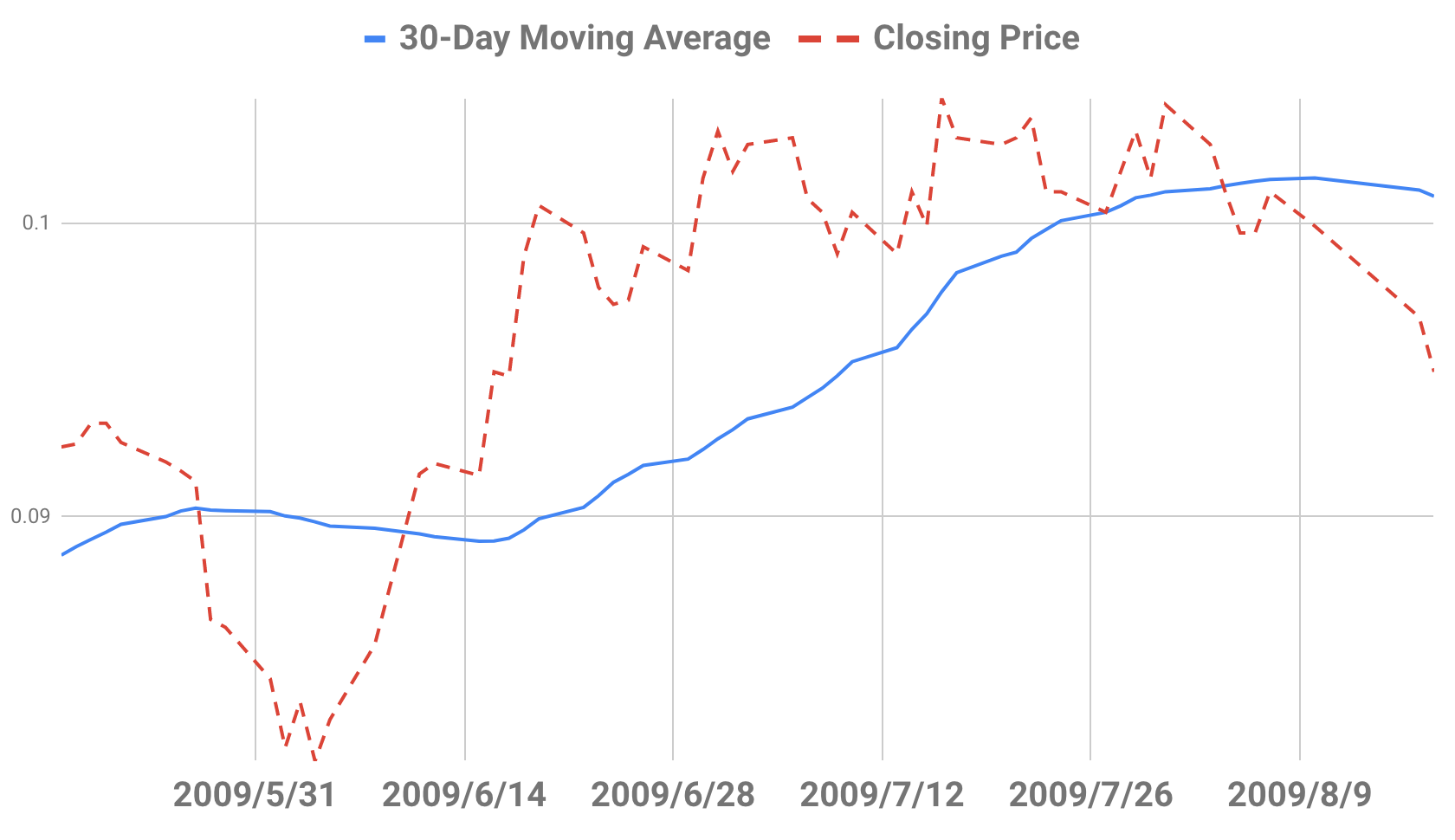} 
        \captionsetup{labelformat=empty}
        \caption{(b) Company B (Company A's customer)}
    \end{minipage}
     \begin{minipage}{0.45\textwidth}
        \centering
        \includegraphics[width=0.9\textwidth]{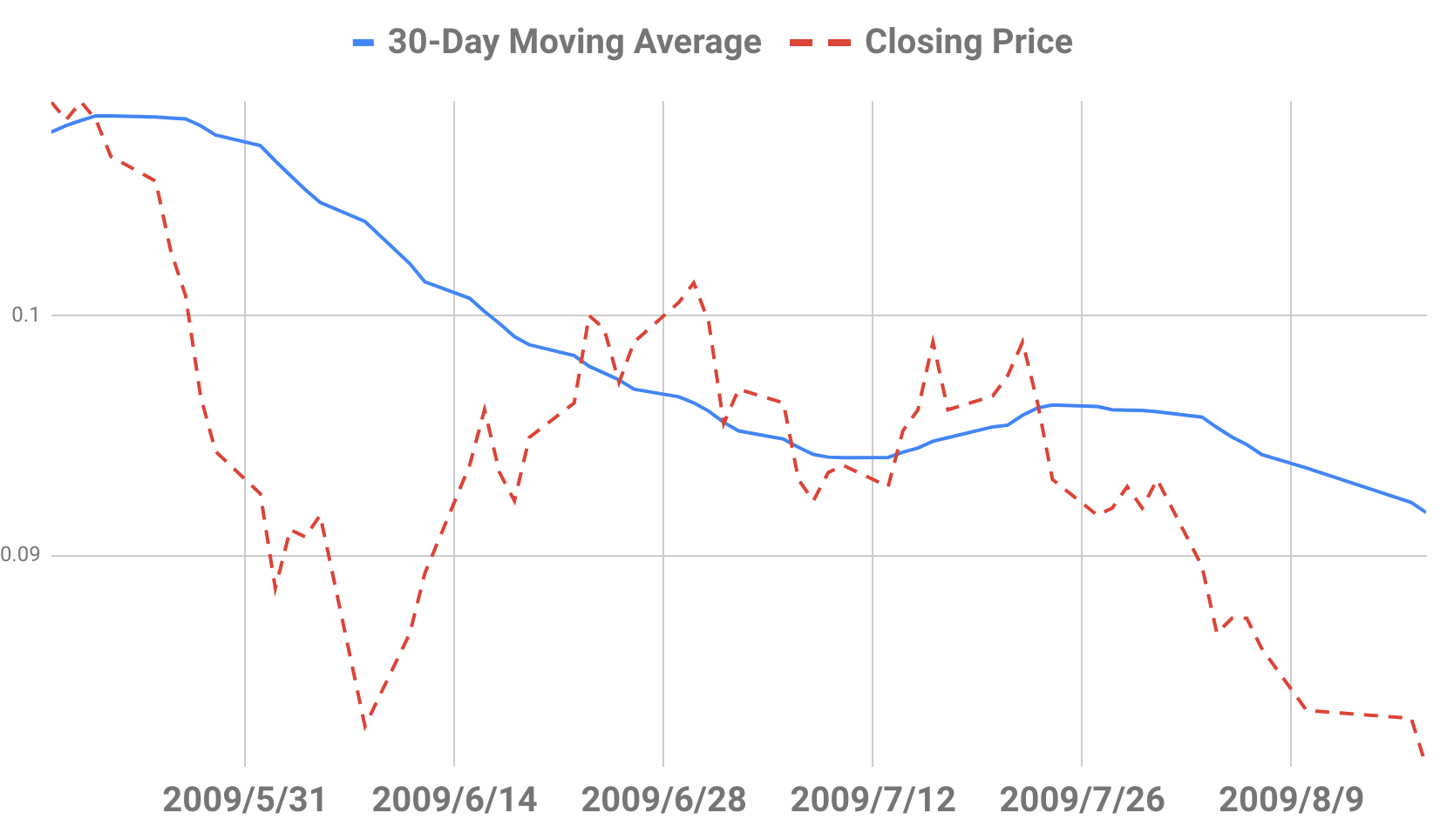} 
        \captionsetup{labelformat=empty}

        \caption{(a) Company A}
    \end{minipage}\hfill
\end{figure}

The nature of the graph data also allows us to investigate post-hoc whether or not related companies have similar stock trends. For example Figure \ref{fig:customer} shows a snapshot of a stock trend chart for two companies of which the model predicted strong edge relations and are connected by customer relations. In this scenario, Company B (Company A's customer) has a trough point on 6/4/2009 and Company A has a similar lagging trough point on 6/8/2009, which suggests that the stock price of a customer contributed to the prediction of stock prices for the target company. It is also important to note that the golden cross point 
has a time lag in a similar fashion. The golden cross occurs at the intersection of the mid-term (30-day) moving average and the closing price, and this indicates a bullish sentiment as the closing price crosses the moving average on an upward trend. As previously mentioned, however, the effectiveness of customer relations don't always hold across all time-spans, and other factors can affect each stock price in different directions and magnitudes. Nonetheless, the intuitive nature of knowledge graph data holds interesting opportunities in terms of explainability of stock predictions.

\section{Possible extensions of current work} \label{discussion}
The scope of this paper was limited to the prediction of 1-day returns for 176 companies listed in the Nikkei 225 market. However, much of the finance industry is concerned with longer prediction timesteps due to the attractiveness of lower transaction fees in passive funds.
Therefore, it can be worthwhile to explore the different relation types with different prediction timesteps to find an effective long-term predictive model. With the current model, the direct customer relation proved to be an effective indicator only for 1-day predictions, where the effectiveness wanes as the timesteps increase, as shown in Table \ref{results-timesteps}. However, it is possible that the accuracy for predictions of longer time horizons will increase as we increase the number of hops. For example, the price of a customer two hops away can be a strong indicator for predicting the 5 or 10-day future returns of the target company, if the correlations propagate. Of course, the prediction of longer time horizons becomes more and more difficult as the timesteps increase due to the multivariate and non-linear nature of the stock market leading to higher error propagation \cite{tensorrnn}. In this scenario, it can also be worthwhile to incorporate other information such as macroeconomic data or news articles rather than solely relying on supply chain relationships, which can be regarded as an important predictor variable in a family of factors which affect the market, rather than the sole indicator. Finally, it is also worthwhile to explore the method of feeding the entire knowledge graph, rather than extracting relations only among the relevant companies listed in the Nikkei 225 market. For instance, if a company X listed in the Nikkei 225 had a relation with an American Company Y, we were only able to connect other listed companies if they also had the same relation with Company Y. This will be possible once we can collect stock price data for Company Y and other non-listed companies.

\section{Conclusion} \label{conclusion}
In this paper, we have conducted preliminary experiments on the effectiveness of incorporating company relations knowledge graphs in predicting the stock market. We have shown that certain relations (customer-of) are effective predictors individually as well as combined with other weaker relations. 
Our preliminary results suggest that knowledge graph data and graph neural networks holds strong promise in creating a more generalizable and practical stock market prediction mechanism. Furthermore, there are many opportunities for extending this work to the field of model interpretability and explinability as the intuitive nature of graph data makes it easier to visualize results. We encourage the finance and AI research community to further explore the use of graph neural networks for the challenging problem of making accurate market predictions and as a consequence, improved investment decisions.
\subsubsection*{Acknowledgments}
We'd like to thank JLB and the GBS Vision Fund (IBM Japan) for supporting this work. We'd also like to thank Yoshiki Minowa, Tomokazu Takeda and Tomohiro Hayashi for many useful discussions and advice.

\medskip

\bibliographystyle{unsrt}
\bibliography{Market_Prediction_KG}

\end{document}